\documentclass[final,hidelinks,onefignum,onetabnum,sort,noabbrev]{siamart220329}
\usepackage[utf8]{inputenc}
\usepackage{bm}
\usepackage[normalem]{ulem}
\usepackage[group-separator={,}]{siunitx}

\usepackage{enumitem}
\usepackage{amssymb}
\usepackage{amsmath}

\usepackage{xcolor}

\usepackage{booktabs}
\usepackage{multirow,multicol}

\usepackage{pgfplots}
\pgfplotsset{compat=1.18}
\usepgfplotslibrary{colormaps}

\usepackage[caption=false]{subfig}

\newcommand{\bfb}{{\bf b}}

\newcommand{\bfe}{{\bf e}}

\newcommand{\bfx}{ {\bf x}}

\newcommand{\bfu}{{\bf u}}

\newcommand{\bfr}{{\bf r}}

\newcommand{\bfv}{{\bf v}}
\newcommand{\bfg}{{\bf g}}

\newcommand{\bfG}{{\bf G}}

\newcommand{\bfK}{{\bf K}}

\newcommand{\bftheta}{{\boldsymbol{\theta}}}

\newcommand{\norm}[1]{\left\lVert#1\right\rVert}

\headers{Implicit Multigrid-Augmented DL for the Helmholtz Equation}{Bar Lerer, Ido Ben-Yair and Eran Treister}

\title{Multigrid-Augmented Deep Learning Preconditioners for the Helmholtz Equation using Compact Implicit Layers\thanks{This research was supported by The Israel Science Foundation grant no. 1589/19 and by the Lynn and William Frankel Center for Computer Science at BGU. IBY is supported by the Kreitman High-Tech scholarship.}}

\author{Bar Lerer\thanks{Computer Science Department, Ben-Gurion University of the Negev, Beer Sheva, Israel. \\
  (\email{barlere@post.bgu.ac.il}, \email{idobeny@post.bgu.ac.il}, \email{erant@cs.bgu.ac.il})\\ $^\ddag$ The authors Bar Lerer and Ido Ben-Yair contributed equally.}
\and Ido Ben-Yair\footnotemark[2] \and Eran Treister\footnotemark[2]{}}

\begin{document}

\maketitle

\begin{abstract}
We present a deep learning-based iterative approach to solve the discrete heterogeneous Helmholtz equation for high wavenumbers.
Combining classical iterative multigrid solvers and convolutional neural networks (CNNs) via preconditioning, we obtain a faster, learned neural solver that scales better than a standard multigrid solver. Our approach offers three main contributions over previous neural methods of this kind. First, we construct a multilevel U-Net-like encoder-solver CNN with an implicit layer on the coarsest grid of the U-Net, where convolution kernels are inverted. This alleviates the field of view problem in CNNs and allows better scalability. Second, we improve upon the previous CNN preconditioner in terms of the number of parameters, computation time, and convergence rates. Third, we propose a multiscale training approach that enables the network to scale to problems of previously unseen dimensions while still maintaining a reasonable training procedure.
Our encoder-solver architecture can be used to generalize over different slowness models of various difficulties and is efficient at solving for many right-hand sides per slowness model. We demonstrate the benefits of our novel architecture with numerical experiments on various heterogeneous two-dimensional problems at high wavenumbers.
\end{abstract}



\section{Introduction}
\label{sec:introduction}
The Helmholtz equation is a partial differential equation \\(PDE) that models the propagation of waves in the frequency domain.
This equation occurs in many disciplines of engineering and science, where its real-world applications include seismic mappings of the earth, magnetic resonance imaging (MRI), optical diffraction tomography (ODT) and more~\cite{bernard2017ultrasonic,soubies2017invodt,guasch2020brainfwi,pham2020lisodt}.
In complex real-world environments, an analytical solution is difficult to obtain.
Thus, numerical methods are typically used, whether based on finite difference discretizations, iterative solvers, or many other approaches.
Indeed, solving the discrete Helmholtz equation efficiently is a substantial field of research~\cite{dwarka2020scalable,gander2019review,graham2020domain,luo2014fast,olson2010smoothed,poulson2013parallel,reps2017additive,sheikh2016accelsl,treister2019multigrid}. 

One common method for the Helmholtz equation is multigrid.
Such methods aim to complement standard local methods called relaxations, which attenuate only high-frequency error components efficiently.
To reduce the low-frequency components of the error, multigrid methods use a solution to the low-frequency error-residual problem on a coarser grid.
However, multigrid alone does not perform well for the Helmholtz equation, mostly due to the indefiniteness of the resulting linear system.

Hence, the shifted Laplacian (SL) approach is often used~\cite{erlangga2006novel,umetani2009multigrid,elman2001krylovenhanced,ERLANGGA2004409}, where the Helmholtz operator is shifted by an imaginary term.
This has the effect of increasing the magnitudes of the eigenvalues of the resulting operator, which leads to better spectral properties of the matrix.
SL works well when used, for example, as a preconditioner to a Krylov method, and a common method for solving SL remains the multigrid approach.

Recently, deep learning has emerged as a capable tool to tackle problems in many disciplines.
Deep learning approaches have been used to solve PDEs as well, with favorable results~\cite{bar2019learning,greenfeld2019learning,moseley2020solving,azulay2021multigridaugmented}. Deep neural networks make use of parameterized combinations of linear operators and point-wise non-linear activation functions, which are stacked into modules often referred to as layers.
Generally speaking, neural networks find a mapping between desired inputs and outputs, where the mapping is ``learned'' by leveraging automatic differentiation and gradient-based optimization to find the parametrization of the network. Such processes rely on large sets of examples and minimize a predefined loss function chosen to correlate with success at the task at hand.
For image-related or grid-like tasks, convolutional neural networks (CNNs) are often used. These are deep neural networks constructed using shift-invariant convolution operations instead of the more general affine maps. Such convolutions are similar to compact stencil-based operators used in numerical solutions of PDEs. 

Neural networks are known as universal approximators, i.e., capable of representing any smooth signal. However, it has been established that to represent highly oscillatory functions, deep neural networks require either substantial depth or other special considerations \cite{rahaman2019spectralbias,tancik2020ffn,li2021fno,sitzmann2020siren}.
Thus, keeping large-scale and highly heterogeneous Helmholtz problems in mind, we expect deep networks to scale poorly with the complexity of the problem.
Successful uses of deep learning for solving highly oscillatory PDEs must therefore find ways to enable the neural networks to deal with the high frequencies inherent to this type of data.

Broadly speaking, three notable strategies have been considered in the deep learning literature for PDEs: physics informed neural networks (PINNs) where neural networks serve as an implicit function that represents the solution to the PDE~\cite{raissi2018hidden,song2021pinnhelm,moseley2020solving,bar2019learning,bar2019unsupervised,han2018solving};
end-to-end methods where networks are trained to consume the parameters of the problem and output its solution~\cite{wiecha2019deep,khoo2018switchnet};
and finally, the hybrid approach where classical methods are augmented with deep networks to solve the problem at hand iteratively.
Recently, works like~\cite{azulay2021multigridaugmented,zhang2022hints,khoo2018switchnet} either adopted existing methods in their architecture or augmented a neural solver with a classical method, e.g., by integrating the neural network as a preconditioner to the classical solver. 
This augmentation approach eases the burden placed on the deep networks since they are no longer required to fully solve the problem on their own.
In turn, this makes the combined solver more robust to the parameters of the problem.

To achieve good performance during training, and enable successful generalization to unseen problems, in this work we exploit the close connection between convolutional neural networks and multigrid cycles.
The two have many commonalities, e.g., both perform convolutions with compact kernels and both up- and downsample the fine-level solution several times.
Therefore, their integration can be achieved seamlessly and efficiently on modern hardware such as graphical processing units.

\subsubsection*{Our Contributions}
In this work, we further explore the use of deep CNNs as accelerators for classical methods, and expand upon~\cite{azulay2021multigridaugmented}, where a convolutional neural network (CNN) is trained to act as a preconditioner to a suitable Krylov solver such as the flexible generalized minimal residual method (FGMRES~\cite{saad1993flexible}). That is, we train an encoder-solver network to cooperate with a classical geometric multigrid solver, the V-cycle, within FGMRES.
Our goal in this work is to improve the scalability and efficiency of the hybrid solver.
Specifically, our contributions are as follows:
\begin{enumerate}[leftmargin=25pt]
\item To improve the efficiency and training time of the CNN, we significantly reduce the number of parameters in the encoder-solver network.
This also reduces the number of floating point operations required in inference at each preconditioning step.
Compared to~\cite{azulay2021multigridaugmented}, we observe no degradation in the performance on the task of solving the heterogeneous Helmholtz equation.
\item We introduce an implicit layer at the coarsest level of the solver U-Net to mimic an exact coarse grid solution, akin to a direct solver typically used at the coarsest level of multigrid methods.
In this implicit layer, a convolution kernel is inverted, rather than multiplied, using FFT~\cite{haber2019imex}.
Initialized as a shifted Laplacian kernel, the kernel weights are optimized during the network training.
\item We adopt a multiscale training method where the training alternates between smaller and larger --- hence more difficult --- problems.
This enables the network to learn salient features quickly in smaller domains, reducing overall training time and cost.
The networks are shown to generalize to larger unseen sizes after being exposed to a smaller number of larger-scale problems.
\end{enumerate}
Lastly, the official implementation for this paper can be found in our GitHub repository\footnote{\url{https://github.com/BGUCompSci/CompactImplicitHelmholtz}}.

\section{Related Work}

\subsection{Physics-Informed Neural Networks}
Several approaches have been proposed to solving differential equations with neural networks.
One such approach is  physics-informed neural networks (PINNs) ~\cite{han2018solving,raissi2019physics,raissi2018hidden,lu2021deepxde}.
In PINNs, the neural network takes on the role of estimating the solution $u$ to a PDE at specific points in space and time, sometimes conditioned on the parameters of the equation.
The differential operators specifying the desired physics are implemented by applying automatic differentiation machinery over the solution network $u$.
Finally, the network is trained using a stochastic optimizer to satisfy the PDE at collocation data points and to minimize a physically-informed residual loss to enforce compliance with the physics being modeled.
Generally, these methods yield a network, often mesh-free, that must be sampled at specific points in space and time to provide the solution of the PDE at those points.
Authors have used this approach for both forward and inverse problems in various fields~\cite{bar2019unsupervised,moseley2020solving,song2021pinnhelm,stanziola2021transcranialhelm}.
Moreover, operator learning approaches have also been proposed, namely~\cite{sirignano2018dgm,lu2021deeponet,li2021fno,patel2021opregression,kissas2022coupledattention}, where the network learns to map elements of the input space directly to the space of solutions, rather than estimate the solution at specific points in the domain.

\subsection{Neural Networks as End-to-end or Iterative Solvers}
Another approach to solving PDEs using neural networks adopts a more traditional approach: optimizing networks with data to solve a given problem by one feed-forward application of the network, or by applying the network iteratively.
Such methods can be viewed as end-to-end methods, since they act as surrogate methods (i.e., in place of a classical solver), that approximate a solution.
However, they suffer from a generalization gap, i.e., the quality of the solutions offered by the networks diminishes the more we need to generalize on. 
For example, generalization could include conditioning on unseen heterogeneous problem parameters.
Generally, since the network acts as the sole solver, the learning task becomes harder the more parameters the solution needs to be conditioned on.
Examples of pure end-to-end approaches include~\cite{wiecha2019deep,xue2020amortizedfem,khoo2018switchnet}.
On the other hand, works like~\cite{hsieh2019learning,pfaff2021meshbased,rizzuti2019learnediterative} apply neural networks iteratively multiple times, improving the obtained approximation of the solution each time.

\subsection{Neural Networks Integrated with Classical Methods}
Finally, authors have recently proposed various ways to integrate deep learning with classical numerical methods.
Often, these approaches speed up classical methods~\cite{um2020solverloop,huang2020intdeep,azulay2021multigridaugmented}, augment weaknesses in end-to-end deep learning solvers~\cite{obiolssales2020cfdnet}, or provide guarantees that deep networks by themselves, do not~\cite{hsieh2019learning,chen2022metamgnet}.
Recent works~\cite{zhang2022hints} integrate neural operator learning with classical methods.
We propose an improved CNN-based preconditioner with an implicit layer, which is applied together with multigrid V-cycles within a classical Krylov solver. 

\section{Preliminaries and Background}
\label{sec:background}

\subsection{Problem Formulation}
\label{sec:helm_problem}

The heterogeneous Helmholtz equation is given by
\begin{equation}
\label{eq:helmholtz}
  -\Delta u(\vec x)-\omega^2\kappa(\vec x)^2(1-\gamma i)u(\vec x)=g(\vec x), \quad  \vec x\in \Omega.
\end{equation}
The unknown $u(\vec x)$ is the Fourier-space representation of the pressure wave function, while $\omega=2\pi f$ denotes the angular frequency of the wave,
$g(\vec x)$ represents any sources present,
$\Delta$ is the Laplacian operator, $i =\sqrt{-1}$, and $\kappa(\vec x)$ denotes the heterogeneous wave slowness model (the reciprocal of the wave's velocity).
The term $\gamma$ indicates the fraction of global attenuation (or damping) in the medium, which is assumed to be very small and constant.
This equation is challenging to solve because it requires finer discretizations as the wavenumber $\omega\kappa$ grows. For large values of $\kappa$, the resulting linear system is highly indefinite~\cite{bayliss1985waveacc,haber2011fasthelm}.
To model open domains, we use an absorbing boundary layer (ABL)~\cite{engquist1977abl,engquist1979radiation,erlangga2006novel}, i.e., a function in $\gamma$ that goes from 0 to 1 towards the boundaries. Sommerfeld, PML~\cite{berenger1994pml,singer2004perfectly} or~\cite{papadimitropoulos2021double} can be viable options as well.

\subsubsection*{Discretization}
Considering a uniform 2D grid of width $h$ in both spatial dimensions, the second-order finite difference discretization of \cref{eq:helmholtz} yields the global linear system
\begin{equation}
\label{eq:aug}
    A^h \bfu^h =\bfg^h,
\end{equation}
where $A^h$ is an operator matrix derived from the stencil
\begin{equation}
\label{eq:ah_matrix}
A^h= -\Delta_h - \omega^2\kappa(\mathbf{x})^2 I = 
    \frac{1}{h^2}
    \begin{bmatrix}
    0 & -1 & 0\\
    -1 & 4-\omega^2 \kappa(\mathbf{x})^2h^2 & -1\\
    0 & -1 & 0
    \end{bmatrix},
\end{equation}
where here we omitted $\gamma$ for simplicity of presentation.
Throughout this work, we denote discrete vectors with boldface letters like $\bfx$.

Typically, at least 10 grid nodes per wavelength are used in a discretization of the Helmholtz equation to obtain an accurate solution.
This means $\omega \kappa h$ is typically bounded by $\omega \kappa h \leq \frac{2\pi}{10} \approx 0.628$.
As a result, the required mesh can be very fine for high wavenumbers, which means that we require significantly more degrees of freedom.
This makes the system \cref{eq:aug} prohibitively large, ill-conditioned, and indefinite.
Due to the boundary conditions, it is also complex-valued.
Moreover, the eigenvalues of $A^h$ in \cref{eq:ah_matrix} will have more negative real parts as the wavenumber $\kappa\omega$ grows.
Thus, solving large-scale systems of this kind often requires the use of efficient iterative solution techniques, like Krylov, multigrid, and other methods.

\subsection{Geometric Multigrid and the Shifted Laplacian}
\label{sec:sl_multigrid}

A common multigrid method to solve the Helmholtz equation is the shifted Laplacian method, originally suggested in~\cite{erlangga2006novel}. Since standard multigrid methods struggle to deal with the indefiniteness of \cref{eq:ah_matrix}, the shifted Laplacian operator
\begin{equation}
\label{eq:hu_sl}
     -\Delta u-\omega^2\kappa(\vec x)^2(\alpha-\beta i)u, \quad \alpha,\beta \in \mathbb{R},
\end{equation}
is used instead in the multigrid solver, and acts as a preconditioner in a suitable Krylov method.  
In this paper we use the pair $\alpha = 1$ and $\beta= 0.5$, which is shown in~\cite{erlangga2006novel} to lead to a good compromise between approximating \cref{eq:aug} and
solving the shifted system using multigrid tools. Specifically, we use a three-level geometric V-cycle and an inexact coarse-grid solution, which we discuss in \cref{sec:GPU}.
The SL multigrid method is very consistent and robust for heterogeneous slowness models, i.e., where $\kappa$ is not uniform.
However, it is considered slow and computationally expensive, especially for high wavenumbers, which is our interest.

Solving PDEs generally requires communicating information between the boundaries and the rest of the domain.
However, if computation relies solely on local operations, information is limited in terms of the distance it can travel.
To facilitate the transmission of information at multiple scales, multigrid methods are commonly used to solve discretized PDEs.
In multigrid, solutions are defined on a hierarchy of grids, where the original fine grid $\Omega^h$ is progressively coarsened.
Two distinct and complementary processes are utilized: relaxation and coarse-grid correction.
Relaxation is done by performing a few iterations of a standard smoother like Jacobi or Gauss-Seidel.
These smoothers have a local nature (e.g., compact convolutions), and hence are only effective at reducing part of the error.
In the case of the Helmholtz system, such relaxation methods do not converge due to the indefiniteness of $A^h$, but one or two iterations of them suffice to smooth the error.
The remaining components of the error typically correspond to eigenvectors of $A^h$ that are associated with small-magnitude eigenvalues, i.e., vectors $\bfe^h$ such that
\begin{equation}
    \norm{A^h\bfe^h}\ll\norm{A^h}\norm{\bfe^h}.
\end{equation}

To reduce these error components, multigrid methods use coarse-grid correction.
The error $\bfe^h$ for an iterate $\bfu^h$ is estimated on a coarser grid, where it is less smooth, and interpolated back to correct $\bfu^h$ on the finer grid.
In other words, we solve an instance of an error-residual equation for $\bfe^H$ projected onto the coarser grid, and then interpolate it back to the fine grid to obtain an approximate $\bfe^h$: 
\begin{equation}
    \label{eq:CGC} A^H\bfe^H=\bfr^H=I^H_h(\mathbf{g}^h-A^h\mathbf{u}^h),\quad \bfe^h= I^h_H\bfe^H.
\end{equation}
The operator $A^H$ approximates $A^h$ on the coarser mesh $\Omega^H$, where $H = 2h$. 

To understand why the indefiniteness of the Helmholtz problem makes multigrid inefficient, consider a smooth error on the fine grid to be a smooth eigen-mode $\bfv^h$ of $A^h$ that corresponds to a small eigenvalue $\lambda^h$. After the coarse grid correction in \cref{eq:CGC}, the new error is approximately~\cite{elman2001krylovenhanced}:
\begin{equation}
\bfe^h = \left( 1 - \frac{\lambda^h}{\lambda^H} \right) \bfv^h,
\end{equation}
where 
$\lambda^H$ is the eigenvalue of $A^H$ that corresponds to the mode in $\bfv^h$ on the coarse grid. Ideally, $\lambda^H$  differs slightly from $\lambda^h$, and both are small in magnitude.
However, if $\lambda^h$ and $\lambda^H$ have opposite signs, the correction is in the wrong direction and will cause the error to increase. This may happen here since $A^h$ is indefinite in our case.

To restrict a fine-grid solution to the coarse grid we use the ``full-weighting'' operator $I_h^H$.
Conversely, to interpolate the coarse-grid solution to a finer grid, we use the bi-linear interpolation operator $I_H^h$.
These operators are defined using the fixed kernels:
\begin{equation}
\label{eq:V_Trans_ops}
    I_h^H = \frac{1}{16}\begin{bmatrix}
    1 & 2 & 1 \\
    2 & 4 & 2 \\
    1 & 2 & 1
    \end{bmatrix},\quad
    I_H^h = \frac{1}{4}\begin{bmatrix}
    1 & 2 & 1 \\
    2 & 4 & 2 \\
    1 & 2 & 1
    \end{bmatrix}.
\end{equation}

Taken as a whole and applied once, the above is the two-grid method, summarized in \Cref{alg:2cycle}. Repeated recursively, this procedure forms a cycle, termed the V-cycle. Note that relaxation is applied twice, before and after the coarse-grid correction, where it is referred to as pre- and post-relaxation, respectively.
This is often done when the coarse system is still too large to solve directly.
The V-cycle is often applied iteratively
to solve the problem to some desired accuracy.

 \begin{algorithm}
 \begin{itemize}[leftmargin=*]
  \item Relax $v_1$ times on $A^h\mathbf{u}^h=\mathbf{g}^h$ with $\mathbf{u}^h$ as an initial guess \;
   \item $\mathbf{f}^{H}\leftarrow I_h^{H}(\mathbf{g}^h - A^h\mathbf{v}^h)$\;
   \item Solve $A^H\mathbf{e}^{H} = \bfr^H$ to obtain $\mathbf{e}^{H}$
   \item $\mathbf{u}^h\leftarrow \mathbf{u}^h + I_{H}^h\mathbf{e}^{H}$\;
  \item Relax $v_2$ times on $A^h\mathbf{u}^h=\mathbf{g}^h$  with $\mathbf{u}^h$ as an initial guess\;
  \end{itemize}
  \caption{Two-grid cycle}
  \label{alg:2cycle}
\end{algorithm}

In the classic V-cycle scheme, one may freely choose the number of levels.
However, unlike other problems, the algebraically smooth error modes of the Helmholtz operator are still quite oscillatory at high wavenumbers. This means very coarse grids typically cannot represent these high-frequency error modes when about 10 grid points per wavelength are used.
Thus, the performance of the solver deteriorates as the number of levels increases.
For example, the results in~\cite{https://doi.org/10.1002/nla.1860} show that three levels achieve the best balance between cost and performance.

\subsection{Convolutional Neural Networks}
\label{sec:cnn}

Neural networks built around shift-invariant convolution operations are called convolutional neural networks (CNNs)~\cite{krizhevsky2012imagenet}.
These CNNs represent some of the most effective methods for dealing with structured high-dimensional data, and for learning spatial features.
Due to the reduction in the number of parameters compared to standard neural architectures, such as the multi-layer perceptron, CNNs are popular for computer vision tasks like image classification, object detection, and semantic segmentation.

We briefly discuss the general structure of CNNs, which primarily employ convolution operations. Thus, a typical CNN is a stack of $N$ steps, or layers, of the form
\begin{equation}
    \label{eq:cnn_layer_example}
    \bfx^{(j+1)} =\sigma \left( \bfK^{(j)} \bfx^{(j)} + \bfb^{(j)} \right), \quad j=0,\ldots,N-1,
\end{equation}
where $\bfx^{(0)}$ is the input signal (often an image), $\bfx^{(j)}$ is known as the $j$-th feature map, which is often a multi-channel image. $\bfK^{(j)}$ denotes a discrete convolution operation with a compactly supported kernel, and $\bfb^{(j)}$ is an optional bias term.
$\sigma$ is an elementwise non-linear function such as the rectified linear unit (ReLU), the sigmoid function or the hyperbolic tangent function.
These functions are known as activation functions, which introduce non-linearity into the structure of the network~\cite{DUBEY202292}.
Usually, the convolutions linearly mix the channels of the input $\bfx^{(j)}$, where a separate kernel is learned for each such input-output channel pair, and each output channel is defined by a linear combination of the convolved input channels. A good guide to popular types of convolutions is given in~\cite{dumoulin2016guide}, of which $3\times3$ is a notable kernel size.

Throughout the layers of a typical CNN, the signal is downsampled while the number of channels is increased. These downsampling operations decrease the size of the signal, and thereby increase the receptive field of subsequent layers. This process also smooths noise from the input~\cite{riad2022learning}. The downsampling is often implemented by layers as given in \cref{eq:cnn_layer_example} but with a ``strided'' convolution, e.g., a convolution with a stride of two that halves the size of the signal, similarly to the restriction operator in \cref{eq:V_Trans_ops}. This results in an encoding process of the input signal, before performing a task such as classification or object detection.

A popular type of CNN is the ResNet architecture~\cite{he2015deep}, which employs blocks of layers given by
\begin{equation}
\label{eq:resnet_layer}
    \bfx^{(j+1)} = \bfx^{(j)} + \bfK_1^{(j)} \sigma \left( \bfK_2^{(j)} \bfx\right), \quad j=0,\ldots,N-1,
\end{equation}
where $\bfx^{(j)}$ and $\bfx^{(j+1)}$ are the input and output features respectively, $\bfK_1^{(j)},\bfK_2^{(j)}$ are two different convolution operators, and $\sigma$ is the non-linear activation function.

Another popular CNN-based architecture is the U-Net~\cite{ronneberger2015unet}, typically used for image-to-image tasks like medical imaging and image semantic segmentation, among others.
The U-Net is made up of two parts, where the first part is an encoder (which involves downsampling and increasing the number of channels), and the latter part is a decoder where the image is upsampled and the number of channels is decreased, until the output resolution is obtained.

\subsection{Implementation Using Graphical Processing Units}
\label{sec:GPU}
Deep neural networks are typically implemented using graphical processing units (GPUs), which are highly parallel computing devices made popular by the emergence of deep learning.
We combine classical solvers with CNNs exploiting the close connection between U-Nets and shifted Laplacian geometric V-cycles.

First, Krylov methods such as GMRES are based on matrix-vector products and thus are quite easy to implement on modern GPU hardware.
The operators in \cref{eq:ah_matrix,eq:hu_sl} are both implemented efficiently as $3\times 3$ convolutions followed by an elementwise vector multiplication for the mass term, taking advantage of the parallelism of GPUs.
In addition, the transfer operators in \cref{eq:V_Trans_ops} are also implemented easily since they are convolution kernels where the stride factor is set to two.
Collectively, all of the operations described here are supported natively in PyTorch~\cite{paszke2019pytorch}, which we use in this work. The convolutions are applied with zero-padding, which is consistent with the Dirichlet boundary condition, whereas the absorbing BCs are obtained through the mass term.
Neumann BCs can be obtained by using reflection or replication padding in CNN frameworks.

On the other hand, the coarsest-grid solution in multigrid is not natural for graphical processors, since direct solvers provide the GPU with fewer opportunities for parallelism, especially when the coarsest grid is large.
To alleviate this, the authors of~\cite{https://doi.org/10.1002/nla.1860} suggest to apply a few iterations of GMRES with a Jacobi preconditioner as the coarsest-grid solver.
We follow this approach here as well, but note that our V-cycle with one pre- and post-relaxation and the iterative coarse-grid solution is cheap and GPU-compatible, but not the most powerful---using other, more serial components such as taking the LU decomposition of the coarsest grid, can result in better approximate solutions.
Still, our goal here remains to accelerate this cheap V-cycle with a convolutional neural network.

\section{Method}
\label{sec:method}

\begin{figure}
    \centering
    \includegraphics[width=\textwidth]{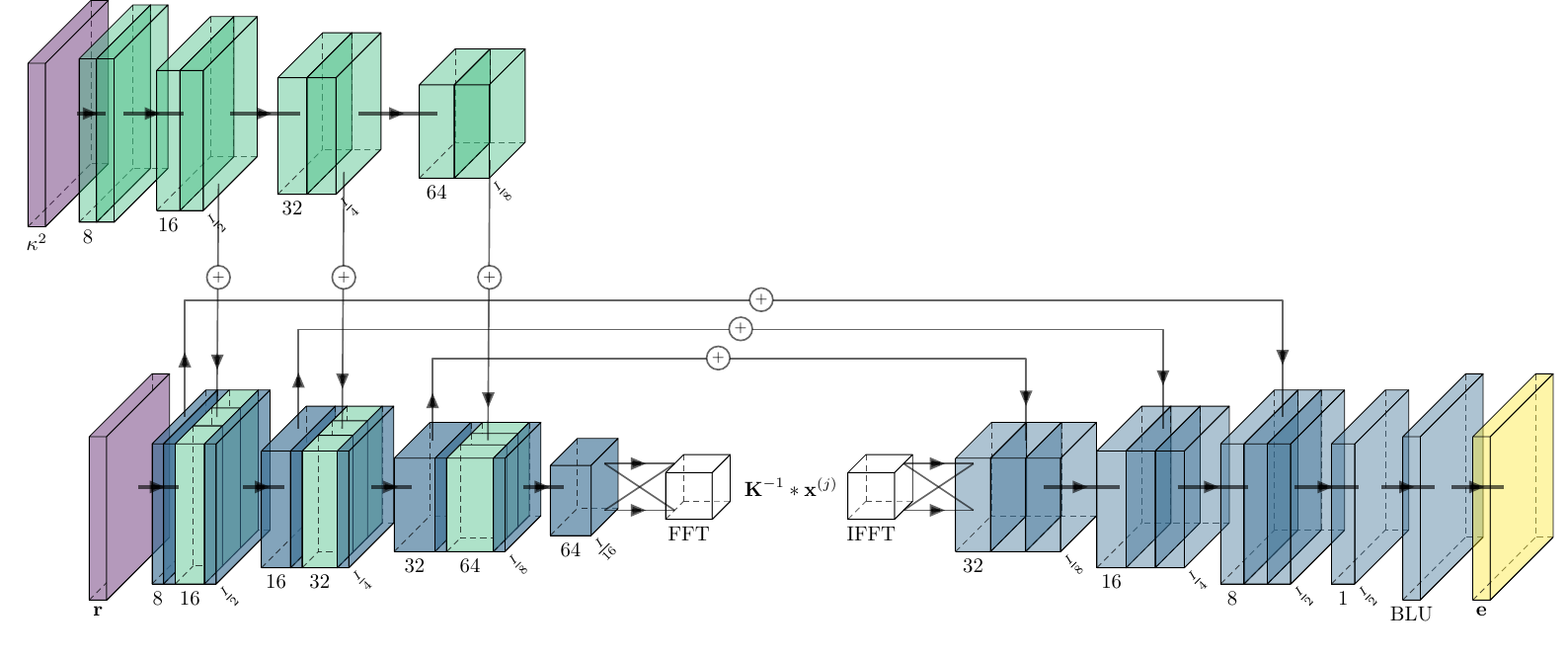}
    \caption{Our implicit encoder-solver CNN architecture.
    The solver network (bottom) maps a residual vector $\bfr$ to an error $\bfe$. The encoder network (top) computes feature maps which are added to the solver architecture as indicated by the arrows in the diagram. Each convolutional block is followed by a batch normalization layer and the softplus activation function.     At the coarsest level of the solver, we use the implicit step defined by \cref{eq:Implicit,eq:Green}.
    Feature maps are then upscaled progressively by learnable convolutional blocks and finally by a non-learnable bilinear upsampling filter (denoted BLU in the figure) back to the original size. }
    \label{fig:implicit_encoder_solver}
\end{figure}

In this section we describe our method, which is based on the U-Net encoder-solver method, proposed in~\cite{azulay2021multigridaugmented}. The U-Net architecture is similar in spirit to the multigrid V-cycle and shares common properties. It consists of a downward section that progressively downsamples the input using convolution operations, and an upward section that upsamples the signal back to its original dimensions. To facilitate the solution of many instances of the Helmholtz equation, we also train an encoder to be applied before the solver network, as we discuss in \cref{sec:EncSol} below. However, the U-Net suffers from a field of view problem as it has a limited number of levels. Hence, we propose a novel implicit layer inspired by solvers modeled after the Lippmann-Schwinger equation, which we present in \cref{sec:lis_layer}. The encoder-solver implicit network is presented in \Cref{fig:implicit_encoder_solver}.  Following that, we present an efficient multiscale training approach in \cref{sec:multiscale_training}, and discuss the improvements in the architecture in sections \cref{sec:lightweight_arch}.
 
\subsection{A Multigrid-Augmented CNN Preconditioner}\label{sec:EncSol} Because the V-cycle converges slowly for high wavenumbers, we augment it using a solver neural network. That is, our preconditioner consists of a forward application of the solver neural network given a batch of residuals as input, generated by a Krylov method. The solver network maps the residuals to corresponding error vectors, which are then fed into a V-cycle for further smoothing before being passed back to the Krylov method. To this end, we train the network on residual and error vectors that are not treated well by V-cycles. In addition, we allow the network to generate noisy predictions to those errors because they can be attenuated well by the V-cycle (i.e., predictions that are generally correct but yield high residuals due to high-frequency noise). This way we simplify the learning task of the CNN, and show that this combination significantly reduces the number of FGMRES iterations required to solve the heterogeneous Helmholtz equation.

The encoder network emits a latent representation of the heterogeneous parameters in $\kappa^2$, which we term the ``encodings'', while the solver network is trained to include the encodings as part of its architecture.
This way, the solver network has more information on $\kappa^2$. To obtain a solution,
the encoder is first applied to the slowness model to generate the encodings;
then, to smooth out the initial error, we start with a few iterations (typically 3) of FGMRES with the V-cycle only as a preconditioner.
The result is used to initialize another invocation of FGMRES, where now the solver network (followed by a V-cycle) acts as a preconditioner. Since the encodings are computed once per slowness model and remain fixed, their cost is amortized over many invocations of the solver network. This is especially desirable when the Helmholtz equation must be solved many times for the same $\kappa^2$, such as when solving inverse problems.
Both the encoder and solver networks are trained together to ensure compatibility.

\subsection{An Implicit Layer based on a Lippmann-Schwinger Solver}
\label{sec:lis_layer}

The U-Net architecture suffers from a field of view problem. This stems from two reasons. First, we train on low resolutions, allowing just a few meaningful levels when training. Second, as we use about 10 grid-points per wavelength, the waves of the Helmholtz solution cannot be represented meaningfully on very coarse levels. These two reasons limit the number of layers we can define in our U-Net. This creates a limited field of view: the U-Net struggles to propagate information over long ranges (e.g., from the boundaries to the center of the domain) when solving a large instance of the Helmholtz problem. Thus, these networks struggle to generalize to higher-resolution domains---especially domains that include more elements than those used to train the U-Net. 

To create a solver that scales better to larger domains we propose replacing the layer at the coarsest level with a novel layer that mimics the exact solution of the coarsest grid in V-cycles, which is obtained using matrix inversion, and has a \emph{global} field of view concerning the spatial space. To this end, we use a variant of the approach of~\cite{haber2019imex} where an ``implicit'' layer includes an inversion of a compact convolution kernel rather than a multiplication with one.

To summarize the approach of~\cite{haber2019imex}, consider \cref{eq:resnet_layer} as a forward Euler discretization of an underlying continuous non-linear ODE or PDE~\cite{ruthotto2020deepmotivated},
\begin{equation}
\label{eq:resnet_layer_cont}
    \partial_t \bfx(t) = f(\bfx(t), \bftheta(t)), \quad t \in [0,T],
\end{equation}
where $f(\bfx(t), \bftheta(t))$ is some non-linear function parameterized by $\bftheta(t)$ and $[0,T]$ is a time interval which is discretized as per the number of layers in the network. The approach of~\cite{haber2019imex} suggested discretizing the time derivative in \cref{eq:resnet_layer_cont} using an implicit (backward) Euler method instead of the explicit (forward) Euler, as in \cref{eq:resnet_layer}. This implicit step is known to be effective in increasing the field of view, but it also requires an inversion of a convolution kernel rather than a multiplication, obtained using the Fast Fourier Transform (FFT)~\cite{haber2019imex}. This is appealing since here essentially we need to invert a spatially dependent kernel in \cref{eq:ah_matrix}. Hence, a natural choice for the CNN solver would be an inverted convolution operator (i.e., implicit), albeit with a fixed (yet learned) kernel. We perform this step only at the coarsest level for several channels ($4\times$ coarsest in each dimension), hence we avoid the high cost associated with running the FFT on the fine feature maps.

However, the exact formulation in~\cite{haber2019imex} did not work well in our setup.
To our understanding, this is probably due to the sensitivity of the network to the boundary conditions, as the FFT considers periodic BCs rather than absorbing ones as we have here. To make our implicit layer suitable for the Helmholtz matrix inversion at hand, we take inspiration from solvers based on the Lippmann-Schwinger (LiS) equation, which has been used to solve inverse problems such as optical diffraction tomography that feature the Helmholtz equation~\cite{soubies2017invodt,pham2020lisodt}.
Essentially, we incorporate this ``classical'' solver as part of our network architecture and allow the network to learn its parameters (i.e., the kernels it inverts). 

Let us first consider the approach of~\cite{soubies2017invodt}, that with~\cite{haber2019imex} inspired the design of our implicit layer. We note that the approach of \cite{soubies2017invodt} is applied in continuous space and then discretized, while here we present the idea on the discrete space. As motivation, consider \cref{eq:ah_matrix}, where we add and subtract the term $\omega^2\kappa_0^2 I$ with a constant $\kappa_0$:
\begin{equation}
\label{eq:LiS1}
A^h(\kappa(\bfx)) = -\Delta_h - \omega^2\kappa^2(\mathbf{x}) I + \omega^2\kappa_0^2 I - \omega^2\kappa_0^2 I.
\end{equation}
Hence, \cref{eq:aug} becomes
\begin{equation}
\label{eq:LiS2}
A^h(\kappa_0)\bfu^h + \omega^2(\kappa_0^2 - \kappa^2(\bfx)) \bfu^h = \bfg^h.
\end{equation}
The approach of~\cite{soubies2017invodt} multiplies \cref{eq:LiS2} by $A^h(\kappa_0)^{-1}$ to form\footnote{The right-hand side $\bfg^h$ in~\cite{soubies2017invodt} is zero.}
\begin{equation}
\label{eq:LiS3}
\bfu^h + A^h(\kappa_0)^{-1}\omega^2(\kappa^2_0 - \kappa^2(\bfx)) \bfu^h = A^h(\kappa_0)^{-1}\bfg^h.
\end{equation}
If $\kappa_0 \approx \kappa(x)$, then we have a well-conditioned system for $\bfu^h$ dominated by an identity matrix which is easy to solve. The matrix $A^h(\kappa_0)^{-1}$ is the inverse of a Helmholtz operator with constant media $\kappa_0$, that can be modeled by a fixed kernel. This inverted shift-invariant operator is equivalent to a convolution with a suitable Green's function sampled on a twice larger grid (at each dimension). This Green's function is convolved at each application of the matrix in \cref{eq:LiS3} using FFT. Because the Green's function is sampled on a large grid, there are no reflections or periodic continuations from the boundaries. In~\cite{soubies2017invodt} the Green's function is defined analytically, and sampled.

We view the process in \cref{eq:LiS2,eq:LiS3} as a preconditioned Helmholtz equation, where the preconditioner is the same operator but with constant media. 
That is, given an approximate solution $\bfu^h$ we approximate the error $\bfe^h$ by
\begin{equation}
    \bfe^h \approx A^h(\kappa_0)^{-1}(\bfg^h-A^h(\kappa(\bfx))\bfu^h) = A^h(\kappa_0)^{-1}\bfr^h.
\end{equation}
This preconditioner has a large field of view and can be implemented in a network efficiently on a GPU using FFT. Let $\widehat\bfG_K$ be the Fourier transform of a Green's function $\bfG_K$ of a kernel $K$ (expanded by a factor of $2$ at each dimension) the implicit layer is given by:
\begin{equation}
\label{eq:Implicit}
\bfx^{(j+1)} = \mathcal{F}^{-1}(\widehat\bfG_K\odot\mathcal{F}(\bfx^{(j)})),
\end{equation}
where $\mathcal{F}$ and $\mathcal{F}^{-1}$ denote the Fourier transform and its inverse, applied per channel, and $\odot$ denotes the elementwise Hadamard product.
More precisely, in the coarsest level of our network, where the number of channels is large but the feature maps are small, we learn a $3\times 3$ kernel to be inverted against each channel.
We train the U-Net to learn the appropriate kernels, which may be viewed as learning the appropriate $\kappa_0$ values for the preconditioner.

To obtain the Green's function of some kernel $K$ in Fourier space needed for \cref{eq:Implicit} (and allow its trainability), we apply a division in Fourier space of the kernel's transform against a point source in the middle of a large domain. That is because division in Fourier space corresponds to matrix inversion using a fixed kernel.
To this end, we first define a point source in the middle of a padded grid (denoted by $\delta$). Then, we take an equally large grid and place the kernel $K$ at the corners of that grid to get $K_{\mathrm{pad}}$ (as described in~\cite{haber2019imex}). We then compute the Green's function by:
\begin{equation}
\label{eq:Green}
\bfG_K = \mathcal{F}^{-1}\left(\frac{\mathcal{F}(K_{\mathrm{pad}})}{\mathcal{F}(K_{\mathrm{pad}})^2 + \epsilon} \odot \mathcal{F}(\delta)\right),
\end{equation}
where $\epsilon=10^{-5}$. The division in \cref{eq:Green} is applied elementwise. The least squares form, obtained by the division by the squared signal, and addition of $\epsilon$ to the denominator are intended to provide numerical stability, since we do not wish to encounter singularities during the learning process, as an arbitrary kernel may represent a singular matrix. To compute $\bfG_K$ using the inverse FFT in \cref{eq:Green} while reducing the influence from the boundaries, we zero-pad the domain to three times the size, and crop it afterward, that is: $\widehat\bfG_K = \mathcal{F}(\mbox{crop}(\bfG_K))$. 

We note that the computation of the Green's functions in Fourier space $\widehat\bfG_K$ is applied per each kernel on the coarsest grid, but only once per batch during training as the kernels' weights do not depend on the input data. Furthermore, $\widehat\bfG_K$ are stored as part of the network and are not computed at all during inference (solve) time. Lastly, since $\delta$ is a fixed point source per resolution, its Fourier transform $\mathcal{F}(\delta)$ is fixed as well and needs to be computed only once per grid for training and inference. 

\subsection{Data Generation and Multiscale Training}
\label{sec:multiscale_training}
Our solver network is intended to work as a preconditioner to a Krylov method, in tandem with a V-cycle.
Therefore, data seen during training must be as similar as possible to the residuals seen during the Krylov method we use (FGMRES).
Consider
\begin{equation}
\bfe^{net} = \mbox{SolverNet}(\bfr,\kappa^2;\theta)
\end{equation}
to be a single forward application of the solver network for a given slowness model $\kappa(\bfx)^2$ and residual $\bfr$.
$\theta$ denotes the set of trainable network weights.
To successfully model a good preconditioner, we seek to minimize the mean squared error (MSE)
\begin{equation}
\label{eq:lossnet}
    \min_\theta \frac{1}{m}\sum_{i=1}^m\|\mbox{SolverNet}(\bfr_i,\kappa_i^2;\theta) - \bfe_i^{true}\|_2^2,
\end{equation}
for each batch of error and residual vectors and slowness models $\{(\bfe_i^{true},\bfr_i,\kappa_i^2)\}_{i=1}^m$, where $\bfr_i=A^h\bfe_i^{true}$.
Since we are dealing with a linear system, where the residual can be computed simply by multiplying the error by the matrix $A^h$, the creation of the aforementioned dataset is straightforward, and there is no need to solve the PDE many times to generate ground-truth solutions. However, since the data must be as close as possible to the residuals seen during runs of the preconditioned FGMRES, we smooth the residuals in the dataset. To this end, we apply a random number of FGMRES iterations with a V-cycle preconditioner (specifically, we apply between 2 to 20 iterations).
That is, starting with a random vector $\bfx_i$, we compute a RHS vector $\bfb_i = A^h \bfx_i$ and apply
\begin{equation}
\label{eq:generatexhat}
  \tilde\bfx_i = \text{FGMRES}(A^h, M=\text{V-cycle},\bfb_i,\bfx^{(0)}=\mathbf{0},iter \in \{2,...,20\}).
\end{equation}
Following that, we compute error vector $\bfe^{true}_i = \bfx_i - \hat{\bfx}_i$ and residual $\bfr_i=\bfb_i - A^h \hat{\bfx}_i = A^h \bfe^{true}_i$ to obtain the $i$-th data sample.
This procedure generates data samples of varying smoothness levels to be used as training residuals. Optimizing \cref{eq:lossnet} against these error-residual pairs directs the network to learn to handle smooth error vectors. The output of the network may be noisy (generating low errors but high residuals), hence in inference time we smooth the network's output using a V-cycle, so it is easy for FGMRES to include it when considering the optimal linear combination of the iterations.

To solve our Helmholtz problem, information must propagate from the boundaries deep into the domain, and vice-versa.
This means that during training, the network must learn to propagate information across the domain regardless of the size of the domain and the location of the boundaries.
Thus, networks exposed only to small domains may struggle to generalize to larger domains.
While CNNs are composed of shift-invariant convolutions, there is a huge influence to the boundaries, especially in multiscale networks like U-Net that reach tiny grids. On the other hand, training on larger domains is expensive.
We show that training the network on multiple scales, i.e., exposing it to varying sizes of problems, enables it to scale better to sizes unseen during training. This especially saves training time compared to training only on the maximal size since most of the iterations are obtained on smaller grid sizes.
To this end, we create datasets of three different sizes: samples are taken from each source dataset, and resized to $128 \times 128, 256 \times 256$ and $512 \times 512$ using bi-linear interpolation. Samples of each respective size are considered a separate dataset for training purposes.
We then alternate between these datasets during training every few epochs (here, 20).
The length of the epochs containing larger training examples is adjusted to be shorter, to use mostly smaller ones, reducing the number of large examples overall.
For example, an epoch of $128 \times 128$ samples is comprised of \num{16000} samples, a $256 \times 256$ is made up of \num{10000} samples and a $512 \times 512$ epoch is only 4000 samples long.
Hence, the training requires less samples of the larger sizes while still performing well.

\paragraph{Slowness Model Datasets}To generate slowness models for training and testing, we use three source datasets of increasing difficulty: CIFAR-10~\cite{Krizhevsky09learningmultiple}, OpenFWI Style-A~\cite{deng2021openfwi} and STL-10~\cite{coates2011stl10}. While CIFAR-10 and STL-10 are natural image datasets and therefore have no bearing on the Helmholtz problem, they serve here as large data sets of general-purpose slowness models to demonstrate our method.

Each source dataset is used to generate three target datasets of three different sizes, as mentioned.
Up to \num{16000} images are sampled from each dataset, resized to the appropriate size, smoothed slightly by a Gaussian kernel, and finally normalized to the range $[0.25, 1]$.
Furthermore, during training only, we shift the imaginary term of $A^h$ by applying a high $\gamma$ value of 0.05. We hypothesize that training on data with a higher $\gamma$ value leads to more consistent training and yields a better convergence rate, due to the domain and boundaries being more absorbent and less reflective. Thus, the network better learns to model wave propagation and generalize to larger domains.
The datasets are then split into training, validation and testing portions for use in training and inference.

\begin{figure*}
    \centering
    \includegraphics[width=0.6\linewidth]{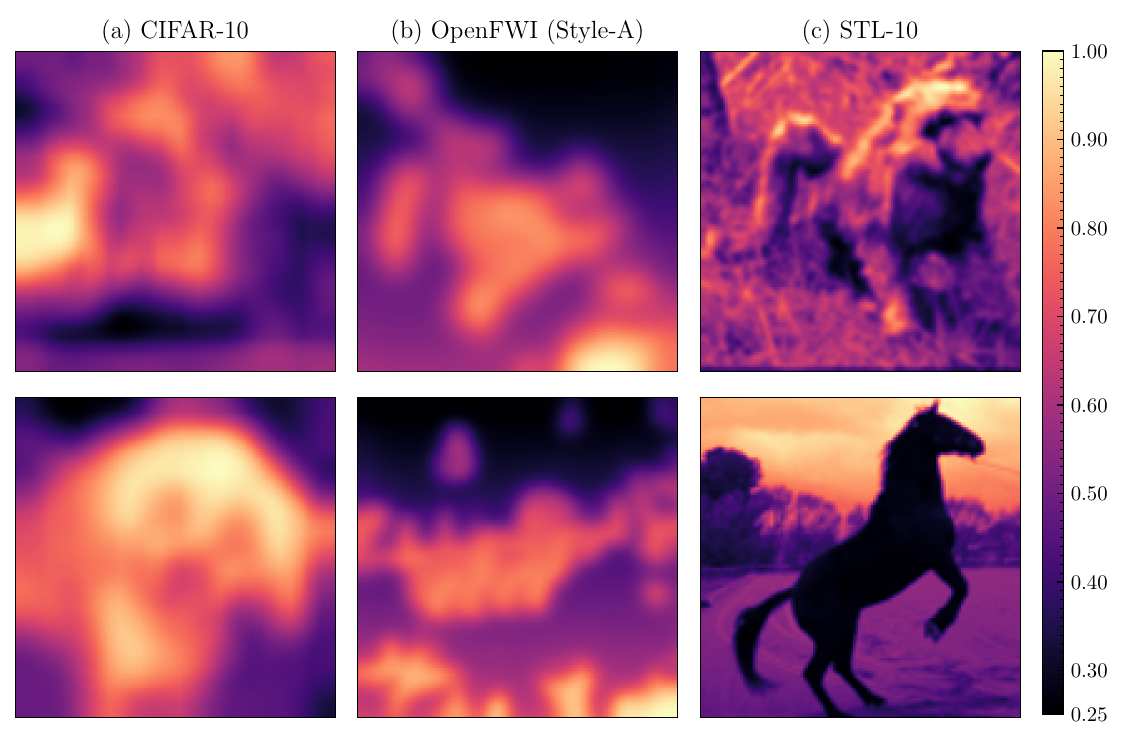}
    \caption{Example slowness models $\kappa^2$ used for training and testing:
    (a) models from the CIFAR-10 dataset;
    (b) models from the OpenFWI Style-A dataset;
    (c) models from the STL-10 dataset.
    We generate separate datasets by sampling up to 16,000 images from each of the datasets.}
    \label{fig:datasets}
\end{figure*}

We note that due to the upscaling, the first two datasets (CIFAR-10 and OpenFWI Style-A) yield significantly smoother, and therefore ``easier'' slowness models.
STL-10, while still smoothed slightly, is still quite challenging.
Example slowness models derived from CIFAR-10, OpenFWI and STL-10 are shown in \cref{fig:datasets}.

\subsection{Lightweight Architecture}
\label{sec:lightweight_arch}
CNN architectures in computer vision have grown over the years, with respect to both the number of parameters and FLOPs associated with the network forward application.
Hence, several patterns have emerged for lighter-weight network designs.
One such technique, credited to MobileNet~\cite{howard2017mobilenets}, uses separable convolutions, where depthwise convolutions and channel-mixing $1 \times 1$ kernels are used separately\footnote{A depthwise convolution is a spatial convolution that is applied on each channel separately with no mixing between the channels. On the other hand, in $1\times 1$ convolutions, there is no spatial operation, and each output channel is a simple linear combination of the input channels.}.
MobileNetV2~\cite{sandler2018mobilenetv2} improves upon the first version by proposing the inverted bottleneck structure, where each network ``module'', similar to \cref{eq:resnet_layer}, consisting of three convolutions: the first is a $1 \times 1$ convolution that expands the number of channels, the second is a depthwise convolution and the last is a $1 \times 1$ convolution that shrinks the number of channels back to the previous smaller number.
This sequence of operations reduces the number of parameters while still making efficient use of the hardware.
We use these techniques in our solver network and in addition, we begin and end the solver network with simple downsampling and upsampling operations  respectively, which are analogous to the MG prolongation and restriction operations in \cref{eq:V_Trans_ops}.
Another method we use to reduce the number of parameters is the use of a sum operation in place of a concatenation operation used in a standard U-Net where feature maps are added together. That is where feature maps computed by the encoder are used in the solver, and in the bypass connections within the solver itself, as shown in \cref{fig:implicit_encoder_solver}. This use of addition instead of concatenation reduces the number of parameters in the network since the resulting number of channels remains unchanged, whereas with concatenation it is doubled.

\section{Numerical Results}

\begin{table}
    \centering
    \caption{Scaling comparison of preconditioner methods.
    Each type of network was trained on problems up to the grid size indicated in the columns to the right.
    Slowness models were taken from the OpenFWI Style-A dataset.
    The average number of FGMRES iterations are shown for a slowness model unseen during training and 1000 right-hand sides, for grid sizes up to $4096 \times 4096$.
    The methods are executed until the residual norm falls by a factor of $10^{-7}$ or 2000 iterations are performed.
    All standard deviations of the CNN-based solvers are less than 10\%, and of the V-cycles are less than 1\%. \\$^\dag$ In these experiments the DL preconditioners did not converge on some instances (less than $8\%$ of the examples), and the standard deviation is less than $15\%$. }
    \small
    \begin{tabular}{l c c c c c c c}
        \toprule
        {\bfseries Data: OpenFWI} & \multicolumn{6}{c}{\bfseries Average number of iterations to solution} \\ 
        \midrule
        {Preconditioner $\setminus$ Grid size:} & $128^2$ & $256^2$ & $512^2$ & $1\text{K}^2$ & $2\text{K}^2$ & $4\text{K}^2$ \\
        \midrule
        Shifted Laplacian V-cycle & 112.9 & 246.8 & 586.9 & 1983.8 & $>2000$ & $>2000$ \\ 
        Explicit U-Net (up to $128^2$) & 18.38 & 30.51 & 52.93 & 178.15 & 404.05$^\dag$ & $>2000$ \\
        Implicit U-Net (up to $128^2$) & 17.26 & 28.98 & 50.63 & 126.09 & 502.34$^\dag$ & $>2000$ \\
        Explicit U-Net (up to $256^2$) & 18.55 & 30.42 & 52.97 & 184.95 & 332.02$^\dag$ & $>2000$ \\
        Implicit U-Net (up to $256^2$) & 17.21 & 28.52 & 47.90 & 92.31 & 281.20$^\dag$ & $>2000$ \\
        Explicit U-Net (up to $512^2$) & 20.53 & 33.29 & 55.31 & 102.41 & 135.74 & 202.96 \\
        Implicit U-Net (up to $512^2$) & 18.82 & 27.34 & 40.62 & 63.39 & 94.13 & 143.03 \\
        \bottomrule
    \end{tabular}
    \label{tab:results_openfwi_scaling}
\end{table}

\begin{table}
    \centering
    \caption{Scaling comparison of preconditioner methods.
    Each type of network was trained on problems up to the grid size indicated in the columns to the right.
    Slowness models were taken from the STL-10 dataset.
    The average number of FGMRES iterations are shown for a slowness model unseen during training and 1000 right-hand sides, for grid sizes up to $4096 \times 4096$.
    The methods are executed until the residual norm falls by a factor of $10^{-7}$ or 2000 iterations are performed.
    All standard deviations of the CNN-based solvers are less than 10\%, and of the V-cycles are less than 1\%.}
    \small
    \begin{tabular}{l c c c c c c c}
        \toprule
        {\bfseries Data: STL-10} & \multicolumn{6}{c}{\bfseries Average number of iterations to solution} \\ 
        \midrule
        {Preconditioner $\setminus$ Grid size:} & $128^2$ & $256^2$ & $512^2$ & $1\text{K}^2$ & $2\text{K}^2$ & $4\text{K}^2$ \\
        \midrule
        Shifted Laplacian V-cycle & 166.2 & 360.72 & 750.93 & 1876.1 & $>2000$ & $>2000$ \\ 
        Explicit U-Net (up to $128^2$) & 25.85 & 44.66 & 75.96 & 233.60 & 284.30 & $>2000$ \\
        Implicit U-Net (up to $128^2$) & 25.46 & 45.19 & 80.67 & 222.85 & $>2000$ & $>2000$ \\
        Explicit U-Net (up to $256^2$) & 28.11 & 34.60 & 54.44 & 77.75 & 139.12 & 280.43 \\
        Implicit U-Net (up to $256^2$) & 27.13 & 33.21 & 48.26 & 69.21 & 132.66 & 232.12 \\
        Explicit U-Net (up to $512^2$) & 27.42 & 34.42 & 54.67 & 77.52 & 155.29 & 231.72 \\
        Implicit U-Net (up to $512^2$) & 26.13 & 33.77 & 47.54 & 63.50 & 130.43 & 189.67 \\
        \bottomrule
    \end{tabular}
    \label{tab:results_stl10_scaling}
\end{table}

In this section, we report our results and demonstrate the efficiency of the proposed techniques, focusing mostly on testing the scalability of the implicit U-Net.
We compare three preconditioning methods: an encoder-solver with an implicit U-Net (as in \cref{fig:implicit_encoder_solver}) followed by a shifted Laplacian V-cycle, an encoder-solver U-Net without the implicit step (called the \emph{explicit U-Net}) also followed by a V-cycle, and lastly a V-cycle-only preconditioner.

We train instances of the explicit and implicit U-Nets, where each is exposed to domains of increasing size in a round-robin fashion: training alternates between $128 \times 128$, through $256 \times 256$ and finally $512 \times 512$.
We train three different network instances for each dataset and network type: one is trained on $128 \times 128$ problems only, the next is trained with samples up to $256 \times 256$, and the last is trained on all three sizes. Each time, we have the same number of total samples. These tests show the influence of the training problem sizes on the test performance on larger problems, showing the importance of multiscale training. We elaborate on the training and architecture details in \cref{sec:arch_training_details}. After training, each network instance, followed by a V-cycle, is tested as a preconditioner to FGMRES(10) on a batch of 1000 right-hand sides. 

The slowness models for training and testing are generated from the CIFAR-10~\cite{Krizhevsky09learningmultiple}, OpenFWI Style-A~\cite{deng2021openfwi} and STL-10~\cite{coates2011stl10} datasets.
We scale each sample up to the appropriate size, smooth it slightly using a Gaussian kernel, and normalize the values to the range $[0.25,1]$, as depicted in \cref{fig:datasets}. 

To demonstrate how the performance of the methods scales with the size of the problem, we measure the average number of preconditioning iterations required to reach a relative residual norm drop of less than $10^{-7}$, or up to 2000 iterations, for domain sizes starting from $128 \times 128$ and up to $4096 \times 4096$.
The results for networks trained on OpenFWI are presented in \cref{tab:results_openfwi_scaling}, and results for STL-10 are shown in \cref{tab:results_stl10_scaling}.
In these tables, note that both the explicit and implicit U-Nets surpass the rather inefficient (for this problem) V-cycle.
Also note that the implicit variant tends to perform better than the explicit variant, and this advantage increases the larger the problems are.
These tables also show that unless trained on the large models of $512^2$, the networks may perform poorly or even fail to converge in larger test problems (e.g., $4\text{K}^2$). This emphasizes the importance of large-scale training in this setting, further motivating the need for the multiscale training in \cref{sec:multiscale_training}. It is clear that the implicit network, trained on larger samples performs the best in the majority of the cases, especially the larger test problems. Additionally, \cref{tab:results_cifar10_scaling} shows a comparison against the results of~\cite{azulay2021multigridaugmented} on the CIFAR-10 dataset, where again, the implicit network that was trained on examples of size up to $512^2$ outperform the other options on the large test cases.

\begin{table}
    \centering
    \caption{Scaling comparison of preconditioner methods.
    Each type of network was trained on problems up to the grid size indicated in the columns to the right.
    Slowness models were taken from the CIFAR-10 dataset.
    The average number of FGMRES iterations are shown for a slowness model unseen during training and 1000 right-hand sides, for grid sizes up to $4096 \times 4096$.
    The methods are executed until the residual norm falls by a factor of $10^{-7}$ or 250 iterations are performed.
    Standard deviations of the CNN-based solvers are less than 10\%, and of the V-cycles are less than 1\%.}
    \small
    \begin{tabular}{l c c c c c c c}
        \toprule
        {\bfseries Data: CIFAR-10} & \multicolumn{6}{c}{\bfseries Average number of iterations to solution} \\ 
        \midrule
        {Preconditioner $\setminus$ Grid size:} & $128^2$ & $256^2$ & $512^2$ & $1\text{K}^2$ & $2\text{K}^2$ & $4\text{K}^2$ \\
        \midrule
        Shifted Laplacian V-cycle & 97.3 & 245.4 & 545.9 & 1964.07 & $>2000$ & $>2000$ \\ 
        Original~\cite{azulay2021multigridaugmented} & 25 & 52 & 101 & N/A & N/A & N/A \\
        Explicit U-Net (up to $128^2$) & 19.10 & 30.61 & 49.64 & 88.19 & 172.02 & 224.51 \\
        Implicit U-Net (up to $128^2$) & 18.12 & 28.35 & 43.89 & 70.84 & 120.12 & 200.41 \\
        Explicit U-Net (up to $256^2$) & 18.25 & 28.96 & 49.47 & 84.40 & 167.18 & 216.75 \\
        Implicit U-Net (up to $256^2$) & 17.21 & 26.91 & 44.98 & 72.43 & 119.95 & 198.16 \\
        Explicit U-Net (up to $512^2$) & 16.76 & 28.73 & 44.51 & 75.32 & 108.33 & 171.15 \\
        Implicit U-Net (up to $512^2$) & 16.75 & 28.03 & 43.29 & 68.08 & 85.20 & 117.29 \\
        \bottomrule
    \end{tabular}
    \label{tab:results_cifar10_scaling}
\end{table}

\subsection{Out-of-distribution Slowness Models}
In addition to the numerical experiments above, we also test our networks on out-of-distribution data, where the distribution of values in the slowness model differs substantially from the data seen during training.
We used the Marmousi~\cite{brougois1990marmousi}, SEG/EAGE Salt-dome, and Overthrust~\cite{aminzadeh1997models} models as out-of-distribution test problems.
These models specify a spatially varying wave velocity $v$, which is inverted to give $\kappa^2 = \frac{1}{v^2}$.
To match the values seen in the training of the networks, we normalized the values such that the maximum of $\kappa^2$ is 1, and took the highest frequency obeying the ten grid-point per wavelength rule. 

\begin{figure}
    \centering
    \includegraphics[width=\textwidth]{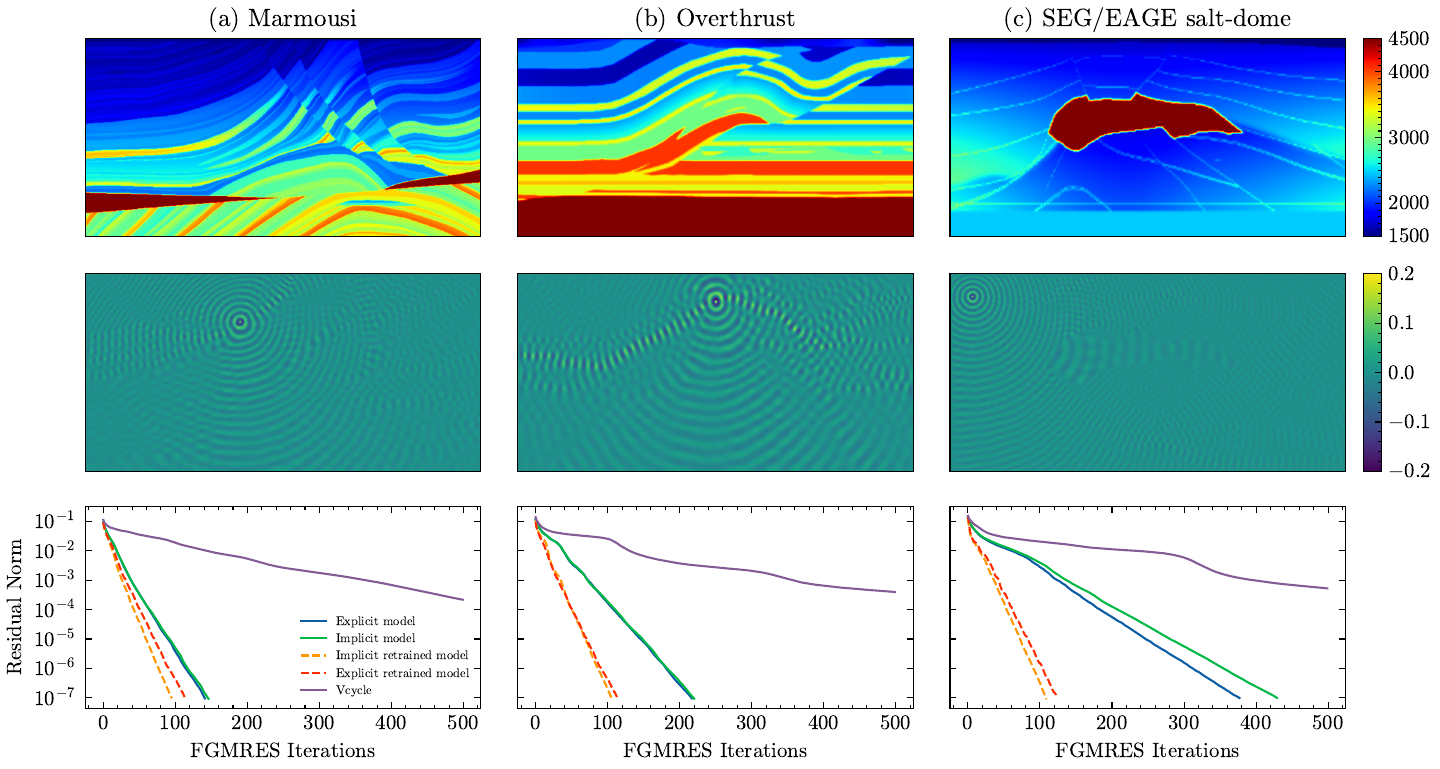}
    \caption{Out-of-distribution test.
    Top: velocity models used for each test~\cite{aminzadeh1997models}.
    Middle: the solution to a single-source Helmholtz problem computed by a single application of FGMRES with the implicit network, followed by a V-cycle, as preconditioner.
    Bottom: convergence plots of the implicit and explicit network preconditioners on each respective problem, as well as a V-cycle-only preconditioner.}
    \label{fig:results_ood}
\end{figure}

\Cref{fig:results_ood} shows how the implicit and explicit networks, trained in earlier experiments, perform on these out-of-distribution models.
Namely, the networks used were trained on $512 \times 512$ domains taken from the STL-10 dataset, i.e. the networks shown in the bottom rows of \cref{tab:results_stl10_scaling}.
For this test, the sizes of the models are $512 \times 1024$ for Marmousi, $352 \times 800$ for Overthrust, and $512 \times 1024$ for the SEG/EAGE Salt-dome model. The pre-trained networks perform reasonably, but the iteration counts somewhat deteriorate compared to the in-distribution iteration counts in \cref{tab:results_stl10_scaling}. To improve these results, we also re-train the CNNs, i.e., optimize them for a short amount of time on the respective out-of-distribution problem.
To save computation, re-training is done on models twice as small as the models used for evaluation, since the retraining is done after the model is known, at solve time.
To this end, we generate 300 pairs of error vectors and their corresponding residuals, and re-train the model for 30 epochs, each comprised of these 300 vector pairs.
Then, we evaluate the networks on new error-residual pairs on the original larger-sized test cases.
\Cref{fig:results_ood} shows how performance on these problems is improved significantly compared to the original networks, at the small cost of additional training. Here, as well, the implicit network has an advantage over the explicit one.

\subsection{Wall-clock Runtime Comparison per Iteration on a GPU}
In our last experiment, we compare the runtime performance of our networks and the other methods considered in this paper.
To this end, the various methods are run as a preconditioner during FGMRES.
The wall-clock time to solution is measured and divided by the number of iterations performed. We report the average time of a preconditioned FGMRES(10) inner iteration, which includes one preconditioning step along with the other actions in the GMRES algorithm.
The experiment is performed 100 times and the results are averaged.
\Cref{tab:results_stl10_runtime} lists the average runtimes per iteration as well as other statistics such as the number of parameters and FLOPs in the network.
The timings do not include the application of the encoder network, which is applied only once per given linear discrete operator (defined by a slowness model $\kappa(\bfx)$), and its output serves all iterations and right-hand-sides with that operator. Hence, its inference time is insignificant compared to the total solution time. For completeness, we report the measures of the encoder network in \Cref{tab:results_encoder_runtime}.
Note that while the implicit U-Net appears to be more expensive than the explicit U-Net, it is likely that a custom implementation (e.g.,~\cite{treister2018low}) of the implicit layer will eliminate much of this performance gap, as the lone implicit layer on the U-Net's coarsest grid is obtained on small feature maps. The results presented here are given for an implementation based on existing PyTorch operations.
Runtimes are listed in \cref{tab:results_stl10_runtime,tab:results_encoder_runtime} for GPU-based as well as CPU-based execution.

\begin{table}
    \setlength{\tabcolsep}{5pt}
    \centering
    \caption{Runtime comparison of solution methods.
    The runtime per preconditioning iteration of FGMRES is averaged over 100 right-hand sides and 100 iterations per RHS.
    FGMRES was run for a random slowness model and for grid sizes up to $4096 \times 4096$ with the preconditioner listed by each respective line.
    Where neural networks are used as preconditioner, they are also followed by an augmenting application of a V-cycle. All standard deviations are too insignificant to list.}
    \begin{tabular}{l c c c c c c}
        \toprule
        {\bfseries Preconditioner} & Params.& FLOPs & \multicolumn{4}{c}{\bfseries Runtime avg. GPU/CPU (s)} \\ 
        {\bfseries Test grid size} & & & $512^2$ & $1\text{K}^2$ & $2\text{K}^2$ & $4\text{K}^2$ \\
        \midrule
        V-cycle only & N/A & 46M & .013/.03 & .01/.15 & .02/.75 & .066/3.13 \\
        Original~\cite{azulay2021multigridaugmented} & 2.5M & 3.14B & .016/.07 & .024/.35 & .056/1.7 & .184/6.8 \\
        Explicit model & 360K & 246M & .015/.042 & .019/.19 & .04/1.03 & .112/4.26 \\
        Implicit model & 360K & 246M & .015/.043 & .019/.19 & .04/1.03 & .114/4.3 \\
        \bottomrule
    \end{tabular}
    \label{tab:results_stl10_runtime}
\end{table}

\begin{table}
    \centering
    \caption{Runtime of encoding slowness models.
    The runtime cost in seconds of applying the encoder to a slowness model of various grid sizes. The cost measurements are averaged over 100 random slowness models. All standard deviations are too insignificant to list.}
    \begin{tabular}{l c c c c c c}
        \toprule
        {\bfseries Encoder} & Params. & FLOPs & \multicolumn{4}{c}{\bfseries Runtime avg. GPU/CPU (s)}  \\ 
        {\bfseries Test grid size} & & & $512^2$ & $1\text{K}^2$ & $2\text{K}^2$ & $4\text{K}^2$ \\
        \midrule
        Original~\cite{azulay2021multigridaugmented} & 1.9M & 5.9B & .003/.07 & .01/.34 & .04/1.56 & .17/5.8 \\
        Ours & 1.2M & 504M & .001/.01 & .003/.07 & .013/.32 & .052/1.3 \\
        \bottomrule
    \end{tabular}
    \label{tab:results_encoder_runtime}
\end{table}

\subsection{Architecture and Training Details}
\label{sec:arch_training_details}
The encoder network is essentially a multi-layer convolutional network that progressively compresses the slowness model, resulting in feature maps of sizes $16 \times \frac{I}{2} \times \frac{I}{2}, 32 \times \frac{I}{4} \times \frac{I}{4}$ and $64 \times \frac{I}{8} \times \frac{I}{8}$, where the first number is the number of channels and $I$ is the original size of the domain along each dimension, e.g., a $2 \times 512 \times 512$ slowness model will be encoded into $16 \times 256 \times 256, 32 \times 128 \times 128$ and $64 \times 64 \times 64$ tensors, respectively.
The domains discussed here are of equal height and width for simplicity, but the design of the network does not rely on this being the case.
The encoding features are computed by a learnable strided convolution followed by two more modules comprised of a learnable convolution, a batch normalization layer, and the softplus activation function.
We term this module the ``downsampling module''.

The solver network is designed to accept these encodings as part of its architecture (see \cref{fig:implicit_encoder_solver}), along with the complex-valued residual vector $\bfr$.
As the computation in the solver network proceeds, the encodings are summed elementwise with the feature maps of the same dimensions.
The results of the sum operation are propagated forward as feature maps.

Our solver network uses four levels, the first three of which are comprised of a learnable downsampling convolution followed by an ``inverted bottleneck'' module, which is discussed in \cref{sec:lightweight_arch}.
This is followed by a single downsampling module, which outputs a feature map of size $64 \times \frac{I}{16} \times \frac{I}{16}$.
For smaller domains (e.g., smaller than $512 \times 512$), we skip the last downsampling operation, and set the last downsampling convolutional layer to instead maintain the size of the incoming feature map (i.e., the coarsest size is maintained at $\frac{I}{8} \times \frac{I}{8}$).
We found that skipping the final downsampling operation is necessary in these cases, since otherwise the resulting feature map is too small for the implicit step to produce meaningful results, causing the solution to diverge (for example, for a $128 \times 128$ domain, the size at the coarsest level would be $8 \times 8$).
Following the implicit step, the feature maps are interpolated back to the original size of the domain by three learnable upsampling modules comprised of a learnable strided transposed convolution followed by two more convolution-normalization-activation modules, with stride set to 1.
Finally, the feature maps are projected back to a single complex-valued channel and upscaled once more by a non-learned bi-linear upsampling filter.
The output of the solver is then the error vector $\bfe$.

\begin{figure}
    \centering
    \includegraphics[width=\textwidth]{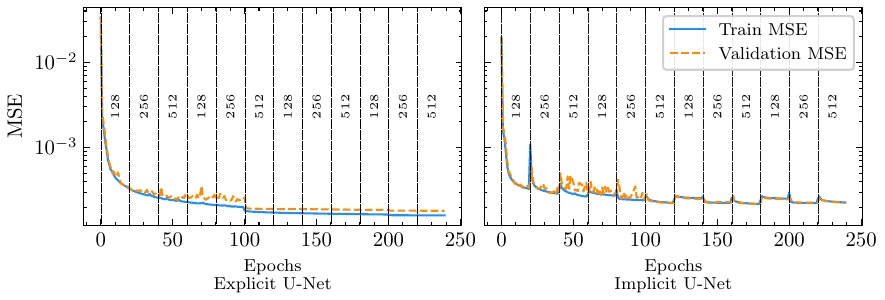}
    \caption{
    MSE loss during training of explicit and implicit U-Net networks with multiscale training on the OpenFWI dataset.
    The MSE loss is as shown in \cref{eq:lossnet}.
    It is worth noting the slight increase in MSE value when the data is switched.
    Our results show that networks trained with multiscale training generalize better to larger unseen sizes.}
    \label{fig:mse_loss}
\end{figure}

We train each encoder-solver pair until convergence on the validation set and up to 250 epochs.
Training was done using the ADAM optimizer~\cite{kingma2017adam} with the default parameters and batch sizes between 30--40, as GPU memory allowed.
The learning rate was initialized to 0.001 and scheduled to divide by 10 every 100 epochs.
For networks trained with more than one dataset (i.e., networks trained on sizes greater than $128 \times 128$), both the training and validation data was switched every 20 epochs in a round-robin fashion.
We used a single current-generation consumer-grade NVIDIA GPU for each training session.
All learnable tensors were initialized randomly using the default Kaiming initialization implemented by PyTorch~\cite{paszke2019pytorch,he2015kaiminginit},
except for the learnable kernel in the implicit layer, which was initialized to $L + iI$ where $L$ is the 5-point Laplacian kernel.
This kernel was then optimized along with the rest of the network's learnable parameters.
Plots of the residual MSEs (as in \cref{eq:lossnet}) under multiscale training are shown in \cref{fig:mse_loss}.

\section{Conclusion}
In this paper, we presented three improvements over the work of~\cite{azulay2021multigridaugmented}.
By introducing our implicit layer to the encoder-solver architecture, we were able to achieve faster convergence and overcome the limited field of view of standard convolutional layers.
Additionally, we adopted architectural enhancements to speed up both the forward and backward computations of the CNNs.

Due to the reduction in the number of parameters and FLOPs, our proposed networks both train and predict faster than previous works, even with the use of the implicit layer at the coarsest level.
Our last contribution in this paper is the use of a multiscale training procedure.
In multiscale training, networks are exposed to problems of different sizes, while larger problems appear less often, thus reducing overall computation time and data demands for training.
In other words, training relies primarily on large numbers of examples of smaller problems, while augmenting with fewer larger problems.

We demonstrated these improvements with supporting numerical experiments on several datasets of increasing difficulty. We showed that networks incorporating the implicit layer lead to better convergence and scaling with the size of the problem.
We also showed that these networks scale better to problems larger than their training data.
Our experiments also show that multiscale training enables better generalization to unseen sizes, compared to networks trained on a smaller variety of sizes. Our architecture was also tested and performed quite well on slowness models that are outside the training data distribution.

Lastly, we presented results that indicate that the forward application of our networks is indeed faster.
For future work, we will consider three-dimensional problems for which the training on reasonable hardware is quite challenging. Furthermore, the performance of both the forward and backward passes could be improved by a custom GPU implementation of these methods (e.g., the implicit layer) instead of relying on readily available PyTorch operators.

\bibliographystyle{siam/siamplain}
\bibliography{refs}

\end{document}